\documentclass[10pt,twocolumn,letterpaper]{article}

\usepackage[pagenumbers]{iccv} 

\definecolor{iccvblue}{rgb}{0.21,0.49,0.74}
\usepackage[pagebackref,breaklinks,colorlinks,allcolors=iccvblue]{hyperref}
\usepackage{amsmath}
\usepackage{amssymb}
\usepackage{mathtools}
\usepackage{amsthm}
\usepackage{lipsum}
\usepackage{enumitem}
\usepackage{csquotes}
\usepackage{flushend}

\def\paperID{14682} 
\def\confName{ICCV}
\def\confYear{2025}

\title{Vision language models are unreliable at trivial spatial cognition}

\author{Sangeet Khemlani \and Tyler Tran \and Nathaniel Gyory \and Anthony M. Harrison \and Wallace E. Lawson \and Ravenna Thielstrom \and Hunter Thompson \and Taaren Singh \and J. Gregory Trafton\\
\\
Navy Center for Applied Research in Artificial Intelligence\\
US Naval Research Laboratory\\
4555 Overlook Dr. SW, Washington, DC 20375, USA\\
{\tt\small sangeet.s.khemlani.civ@us.navy.mil}
}

\begin{document}
\maketitle

\begin{abstract}
Vision language models (VLMs) are designed to extract relevant visuospatial information from images. Some research suggests that VLMs can exhibit humanlike scene understanding, while other investigations reveal difficulties in their ability to process relational information. To achieve widespread applicability, VLMs must perform reliably, yielding comparable competence across a wide variety of related tasks. We sought to test how reliable these architectures are at engaging in trivial spatial cognition, e.g., recognizing whether one object is left of another in an uncluttered scene. We developed a benchmark dataset -- \textsc{TableTest} -- whose images depict 3D scenes of objects arranged on a table, and used it to evaluate state-of-the-art VLMs. Results show that performance could be degraded by minor variations of prompts that use logically equivalent descriptions. These analyses suggest limitations in how VLMs may reason about spatial relations in real-world applications. They also reveal novel opportunities for bolstering image caption corpora for more efficient training and testing.
\end{abstract}

\section{Introduction}
\label{introduction}

The human ability to process spatial information is flexible, consistent, and productive. For instance, an able-minded adult who understands the information expressed by the sentence, \textit{the car is to the left of the truck}, can respond accurately to a variety of verbal prompts, such as:

\begin{itemize}[nosep]
  \item \textit{what is to the right of the truck?}
  \item \textit{how are the car and the truck positioned?}
  \item \textit{is the truck to the left of the car?}
\end{itemize}

\noindent and so on. Vision-language models (VLMs) are AI systems that couple language models with vision encoders; they are trained on vast corpora of image-text pairs. Breakthroughs in their development have demonstrated possible humanlike performance on a large swath of visual tasks such as image tagging and automatic captioning \cite{hochmair2024correctness, zhang2024vision}, leading some researchers to argue that VLMs can exhibit robust spatial scene understanding \cite{cai2024spatialbot, fu2024scene, chen2024spatialvlm, cheng2024spatialrgpt, ghosh2024exploring, chen2024large}. Others argue that current architectures are susceptible to unique forms of attack \cite{liu2024survey} and hallucination \cite{liu2024survey2} and cannot process spatial imagery as capably as humans \cite{liu2023visual}. 

For instance, one recent benchmarking study \cite{kamath2023s} showed that state-of-the-art VLMs had difficulty processing relational information in consistent ways. The authors developed a database of real-world images of, e.g., a mug and a plate in various positions, and they tested whether VLMs could select an appropriate description from a set of four alternatives (e.g., ``A mug is \textit{in front of} a plate'', ``A mug is \textit{behind} a plate'', and so on). The best performing VLM (BLIP-VQA) correctly identified relations for 48\% of the benchmark images, well below estimated human levels of performance. The authors proposed various explanations for why VLMs struggle at this task: one compelling explanation is that prepositions such as ``left of'' and ``right of'' occur very rarely in the large scale image-caption corpora that are often used in training models. Another is that prepositions can sometimes be ambiguous, e.g., ``the car is to the left of the truck'' could be true from a viewer's (egocentric) frame of reference, or it could be true from the object's frame of reference, such that the driver's side of the car denotes its left and the passenger's side denotes its right (in the US). Hence, any picture that shows a car next to a truck could be sensibly labeled with that caption, which can distort how VLMs learn relations. The authors propose various ways of improving VLM performance.

Researchers have conducted other sorts of surveys on how VLM perform in scene understanding tasks. Some studies reveal ways of improving performance \cite{tang2024sparkle, nagar2024zero, yuksekgonul2023}, while others reveal systematically low performance on synthetic data \cite{zhang2024vision2}. Yet studies seldom investigate the reliability of VLM spatial capacities across comparable prompts and tasks. Task reliability is necessary for widespread VLM applicability, because if VLMs operate reliably across comparable tasks, then, even if they perform worse than humans, those deficits can be anticipated and mitigated systematically. If, in contrast, the same VLM performs sensibly on one task but poorly on a minor variation of that same task, then its overall performance is unpredictable for practical application.

Consider the evaluation prompt used by \cite{kamath2023s}, which queried VLMs to select one of four sentences that best described the image. Suppose the authors had used an alternative prompt, identical in all respects to the original except that it inverted the order of the objects (e.g., ``A \textit{plate} is behind a \textit{mug}''). A VLM that performs reliably should yield identical performance for these two prompts, since they are logically equivalent. These surface-level variations should not perturb any underlying capacity for perceiving equivalent relations. Suppose further that the researchers used a different prompt altogether, one in which the VLM had to indicate whether each of the four sentences was true or false. Here, too, the prompt and output of the VLM may be distinct (the VLM should produce a vector of four truth values) but its ability to detect relational information should be comparable to the prompt used by \cite{kamath2023s} and its alternative.

We therefore conducted evaluations on whether VLMs exhibit reliable spatial scene understanding on ``trivial'' tasks, that is, those on which humans should perform easily, quickly, and accurately. We benchmarked performance using a synthetic dataset -- \textsc{TableTest} -- which is composed of images that depict 3D scenes of one, two, or three objects arranged side-by-side on a table using objects selected from a standardized database, Objaverse \cite{deitke2023objaverse}. We evaluated state-of-the-art VLMs on their ability to comprehend distinct but related prompts of rudimentary relations (\textit{left} and \textit{right}) with imagery and with text alone. Results reveal poor reliability across the separate prompts, and more critically, across variations of prompts that should have no affect on performance.

In what follows, we describe:
\begin{itemize}[nosep]
  \item why performance on trivial cognitive tasks is a better benchmark of VLM performance than more complex reasoning tasks;
  \item why the synthetic dataset we developed is useful for both testing and training spatial reasoning;
  \item why our investigations suggest systematic limitations in how VLMs construct and process spatial relations;
  \item how the use of minor prompt variations could help make more efficient use of training data.
\end{itemize}

\noindent Unless addressed directly, these limitations may prohibit widespread VLM adoption.

\subsection{What constitutes ``trivial'' spatial cognition?}

We describe any behavior as \textit{trivial} if it is reasonable to expect the vast majority of neurotypical adults to produce it. Trivial forms of spatial cognition include:

\begin{itemize}[nosep]
  \item Consistent usage of \textbf{spatial demonstratives} such as ``this'' and ``there'', e.g., using ``this'' to refer to proximal context and ``there'' to refer to distal context \cite{coventry2023spatial, landau1993whence}
  \item Rudimentary \textbf{spatial reasoning}, e.g., understanding that the sentence \emph{A is to the left of B} implies that \emph{B is to the right of A} \cite{knauff2013space, byrne1989spatial}
  \item The \textbf{perception and recognition} of primitive relations (e.g., \emph{left}, \emph{right}, \emph{above}, \emph{on}, \emph{inside of}, \emph{in front of}, \emph{next to}, \emph{between}) in uncluttered scenes \cite{hayward1995spatial}
\end{itemize}

\noindent These behaviors serve as precursors to conscious and deliberative spatial reasoning \cite{ragni2013theory, cortes2021makes}. For example, complex inferences about the stability of objects in a scene require initial awareness of the relations within it \cite{ullman2018learning}. 

Assessing trivial spatial competency in VLMs is an efficient way of gauging their capacity to cope with more complex tasks. If systems are incapable and unreliable at producing trivial spatial behaviors, they have little hope of handling nontrivial tasks, such as detecting and maintaining coherence across large bodies of visual and text data. The study of trivial spatial cognition also permits the generation of synthetic data for testing and evaluation. In particular, the detection of spatial relations in images can serve as an ideal candidate for training and testing spatial competence in VLMs, because it is straightforward to generate and manipulate imagery to reflect specific relational information.

\section{\textsc{TableTest}: A benchmark for spatial relation recognition}

\begin{figure}[!t]
\vskip 0.2in
\begin{center}
\centerline{\includegraphics[scale=0.23]{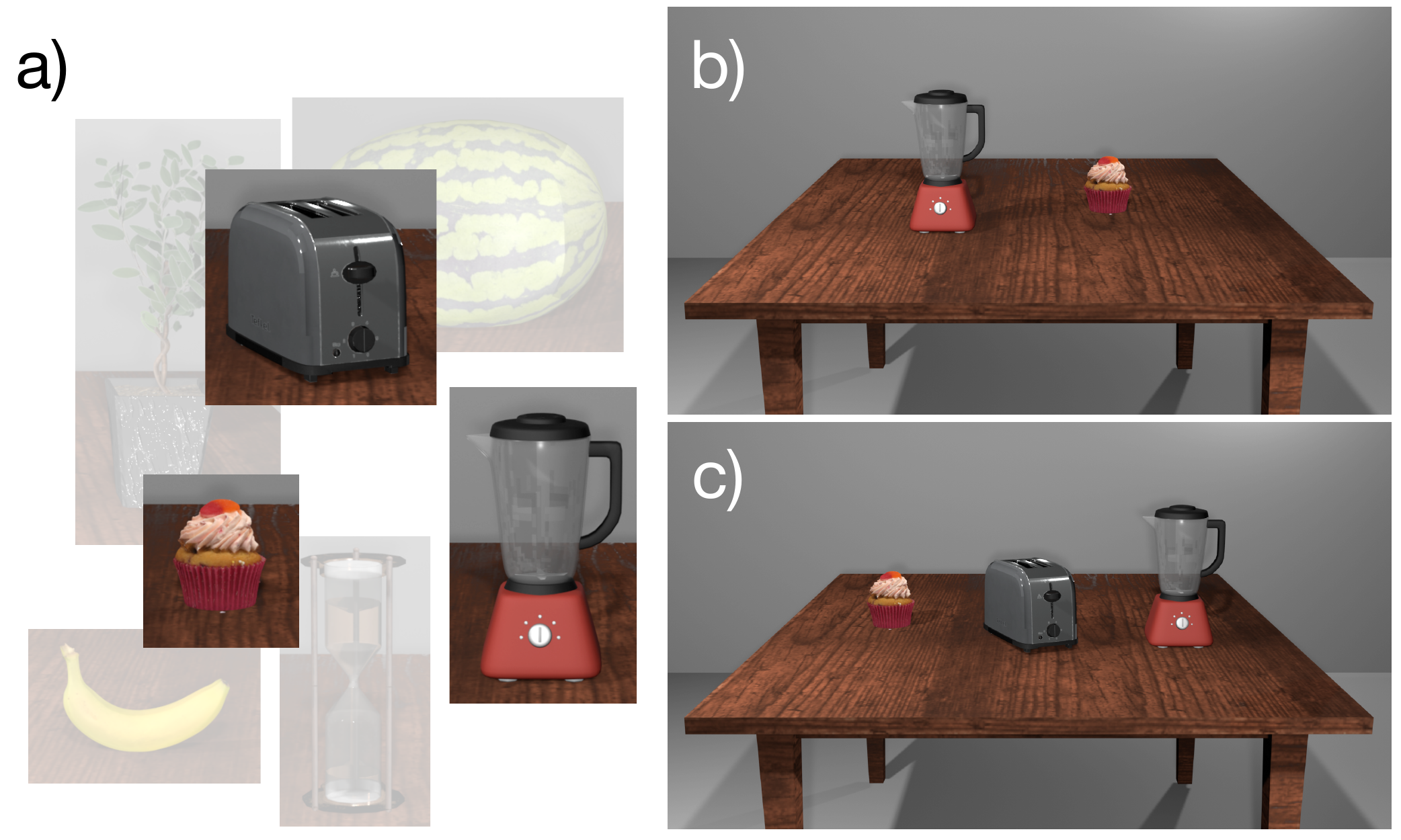}}
\caption{Examples from the \textsc{TableTest} dataset, which includes 64 individual objects (a) in various configurations (b, c). The dataset includes all 2-object configurations, such as one in which a blender is depicted to the right of a cupcake (b). It also includes all 3-object configurations, such as one in which a cupcake is to the left of a toaster, which is to the left of a blender (c).}
\label{figure-tabletest}
\end{center}
\vskip -0.2in
\end{figure}

We developed \textsc{TableTest}, a synthetic dataset designed to test spatial relation recognition and reasoning in VLMs. Each image in the dataset depicts one or more objects arranged on a table (see Figure \ref{figure-tabletest}). We selected 64 objects from the Objaverse dataset of annotated 3D objects \cite{deitke2023objaverse} on the basis of the following criteria:

\begin{itemize}[nosep]
  \item Selected objects had limited association with one another (e.g., \textit{blender} and \textit{screwdriver}) instead of all coming from a given environment (kitchen appliances).
  \item All objects were small enough to sensibly rest on one portion of a table, but nevertheless varied in size, material, color, and texture.
  \item Objects included both natural items (\textit{tomato}, \textit{banana}) as well as artifact kinds (\textit{telephone}, \textit{lamp}).
\end{itemize}

\noindent We hand-scaled these objects to ensure plausible relative object sizes, and used Blender 3D (\texttt{www.blender.org}) to ensure uniform table placement, such that the center point of each object was placed 200px away from the center point of the table in the left or right directions. \textsc{TableTest} consists of all possible 2- and 3-object configurations of the 64 objects, yielding a dataset of 4,032 2-object images and ~250K 3-object images. We used \textsc{TableTest} to conduct evaluation tests of spatial relation recognition behavior in readily available VLMs.

\begin{table*}[!h]
\centering
\begin{tabular}{p{1.5cm}p{6cm}p{8cm}}
\hline
 \textbf{Prompt} & \textbf{Template} & \textbf{Prompt variations/options} \\ 
\hline
 1 & ``Is the following statement true or false: the [object] is to the [relation] of the [object]?'' & \begin{tabular}{@{}l@{}l@{}l@{}}\textit{the A is to the right of the B} [relation is false]\\\textit{the B is to the right of the A} [relation is true]\\\textit{the A is to the left of the B} [relation is true]\\\textit{the B is to the left of the A} [relation is false]\end{tabular} \\ 
 2 & ``Please select a correct relationship from:'' [4 options] & \begin{tabular}{@{}l@{}l@{}l@{}}\\\textit{The A is to the right of the B}\\\textit{The A is to the left of the B}\\\textit{The B is to the right of the A}\\\textit{The B is to the left of the A}\end{tabular} \\  
 3 & ``Please select a relationship that does not hold from:'' [4 options] & \begin{tabular}{@{}l@{}l@{}l@{}}\\\textit{The A is to the right of the B}\\\textit{The A is to the left of the B}\\\textit{The B is to the right of the A}\\\textit{The B is to the left of the A}\end{tabular}\\ 
 4 & ``Is the [object] [disjunction of relations] the [object]?'' & \begin{tabular}{@{}l@{}l@{}l@{}}\\\textit{Is the A to the left of or to the right of the B?}\\\textit{Is the A to the right of or to the left of the B?}\\\textit{Is the B to the left of or to the right of the A?}\\\textit{Is the B to the right of or to the left of the A?}\end{tabular}\\ 
 5 & ``Is the [object] [disjunction of relations] the [object]?'' & \begin{tabular}{@{}l@{}l@{}l@{}}\\\textit{Is the A inside of, to the left of, or to the right of the B?}\\\textit{Is the A to the right of, to the left of, or inside of the B?}\\\textit{Is the B inside of, to the left of, or to the right of the A?}\\\textit{Is the B to the right of, to the left of, or inside of the A?}\end{tabular}\\ 
 6 & ``Fill in both blanks according to the image:'' & \begin{tabular}{@{}l@{}}\\\textit{The [blank] is to the left of the [blank]}\\\textit{The [blank] is to the right of the [blank]}\end{tabular}\\ 
 7 & ``Fill in the blank according to the image:'' & \begin{tabular}{@{}l@{}}\\\textit{The [blank] is on the left side of the table.}\\\textit{The [blank] is on the right side of the table.}\end{tabular}\\ 
 8 & ``Fill in the blanks according to the image:'' & \begin{tabular}{@{}l@{}}\\\textit{The A is to the [blank] of the B}\\\textit{The B is to the [blank] of the A}\end{tabular}\\ 
\hline
\end{tabular}
\caption{Eight different types of prompts for testing 1D spatial relation recognition performance in vision language models. Each prompt corresponds to a distinct cognitive task that humans can perform with ease, e.g., Prompt 1 is a true-or-false question, Prompts 4 and 5 are yes/no questions, and so on. Each prompt can vary along numerous structural and conceptual dimensions, e.g., Prompt 1 can probe whether the queried relation is \textit{left} or \textit{right}, and it can probe whether the relation is true or false given additional text or an input image. Variations should have no effect on VLM performance. The table lists the variations and/or options provided to various VLMs in the evaluation studies described in the text.}
\label{table-prompts}
\end{table*}

\section{Evaluating trivial spatial relation recognition in VLMs}

Spatial relation recognition behavior should be accurate and consistent across different kinds of verbal prompts, such as those provided in the introductory paragraph above. To test the capacity of VLM architectures to cope with relation recognition, we subjected common VLMs to a battery of prompts that varied in task and format. We included VLM architectures for evaluation based on three criteria: they were recently released (post-2022), freely available and well documented, and capable of single-shot identification of the 64 objects in \textsc{TableTest} without additional training. We identified three architectures that matched those criteria: Idefics2 \cite{laurenccon2024matters} (8B parameters), InstructBlip-Vicuna (7B parameters), which builds atop the BLIP architecture \cite{li2022blip}, and Llama 3.2 \cite{touvron2023llama} (11B parameters). Initial testing verified that all three VLMs identified the 64 single-object images in the dataset at or above 96\% accuracy, and we measured accuracy in a manner that was lenient to systematic generation of synonyms and misidentification (see Appendix \ref{appendixA}).

\subsection{Evaluation methodology}

Evaluation of spatial relation recognition consisted of two phases. The first phase examined multimodal input: we subjected each two-object image in \textsc{TableTest} to prompts that queried for spatial relations depicted in the image (see Table \ref{table-prompts}). A second comparison phase used text-only translations of those same \textsc{TableTest} images to run analogous queries. For example, a text-only version of Figure \ref{figure-tabletest}b is: 

\label{text-only-example}

\begin{displayquote}
{\small{\texttt{The blender is on the left side of the table. The cupcake is on the right side of the same table.}}}
\end{displayquote}

\noindent Prompts (and image descriptions) were minimal in nature to permit systematic comparison (see Table \ref{table-prompts}). For example, Prompt 4 asked for yes/no answers to questions such as ``Is the blender to the left of the cupcake?'' Additional instructions could help to regiment output and would permit easier analysis of VLM performance -- but they could also introduce biases and confounds, and they could make it more difficult to compare one type of prompt to another, since different prompts would require different sorts of instruction. We therefore opted for prompts that were concise and immediately interpretable to a lay audience. An advantage of this approach was that each prompt could be used to identify relations from either multimodal or text-only input. 

We likewise devised surface-level structural variations of the different prompts to test for uniformity in VLM responses. Each of these variations should have no effect on VLM relational recognition. For example, we asked 4 separate versions of Prompt 1 by varying a) whether the prompt used \emph{right} or \emph{left} in its description, and whether the image depicted the relation or not (see Table \ref{table-prompts}). Hence, a VLM that received Prompt 1 paired with Figure \ref{figure-tabletest}b should respond {\small{\textbf{\texttt{false}}}} when asked to judge whether it's true or false that \emph{the cupcake is to the left of the blender} or \emph{the blender is to the right of the cupcake} .

As we show below, VLM performance across different prompts reveals unreliable performance for both multimodal and text-only evaluations.

\subsection{Overall performance}

Since each query concerned a trivial task for a VLM to perform, adequate performance should be $\geq$ 90\% accuracy for every prompt and prompt variation under investigation. Table \ref{table-overallperformance} shows overall proportions of accuracy across spatial relation recognition prompts for BLIP, Llama3, and Idefics2 for both multimodal and text-only evaluations. Figures \ref{figure-multimodal} and \ref{figure-textonly} plot accuracy distributions for multimodal and text-only evaluations respectively. Overall accuracies across 8 different prompts ranged from .12 to 1.00. Ranges for the different VLMs were as follows: BLIP [.12, .96]; Idefics2 [.20, 1.00]; Llama3 [.12, .98].

Some prompts yielded high performance in a text-only context (e.g. Prompt 7) but low performance in a multimodal context; other prompts yielded the opposite pattern (e.g., Prompt 5). Each VLM yielded inadequate performance ($\leq$ 90\% accuracy) on two or more prompts for both multimodal and text-only contexts. VLMs reached adequate performance on 3 of the 8 prompts used in multimodal contexts and 7 of 8 prompts used in text-only contexts. For multimodal contexts, Llama3 outperformed other models on 5 prompts; Idefics2 outperformed other models on 3 prompts; and BLIP consistently underperformed relative to other models. For text-only contexts, Idefics2 outperformed other models on 6 of 8 prompts; Llama3 outperformed other models on 2 of 8 prompts; and BLIP underperformed.

\begin{table}
\centering
\resizebox{\columnwidth}{!}{\begin{tabular}{rcccccccc} 
\hline
\multicolumn{1}{l}{} & \multicolumn{8}{c}{\textbf{Prompt}}                                                                           \\ 
\cline{2-9}
\multicolumn{1}{l}{} & 1            & 2            & 3            & 4            & 5            & 6            & 7            & 8             \\ 
\hline
\multicolumn{1}{l}{} & \multicolumn{8}{c}{\textit{Multimodal}}                                                                       \\
BLIP                 & .50          & .16          & .38          & .51          & .31          & .36          & .47          & .52           \\
Llama3               & .86          & .86          & .44          & \textbf{.98} & \textbf{.97} & .26          & .88          & .74                 \\
Idefics2             & \textbf{.95} & .84          & .43          & .85          & \textbf{.90} & .39          & .89          & .84        \\
                     &              &              &              &              &              &              &              &                     \\
\multicolumn{1}{l}{} & \multicolumn{8}{c}{\textit{Text-only}}                                                                                       \\
BLIP                 & .82           & .61           & .36          & .72           & .67          & .75           & \textbf{.96}  & \textbf{.93}    \\
Llama3               & \textbf{.99}  & \textbf{.93}  & .85          & \textbf{.98}  & .87          & \textbf{.96}  & \textbf{.98}  & \textbf{.93}            \\
Idefics2             & \textbf{.99}  & \textbf{.95}  & .86          & \textbf{1.00} & \textbf{.99} & .88           & \textbf{.97}  & \textbf{1.00}  \\
\hline
\end{tabular}}
\caption{Overall proportions of accuracy for VLM performance on trivial spatial recognition prompts (see Table \ref{table-prompts}) based on multimodal and text-only evaluations. Bolded cells highlight acceptable performance ($\geq$ .90) for the  queries used in each evaluation.}
\label{table-overallperformance}
\end{table}

\subsection{Multimodal context evaluations}

\begin{figure*}[ht]
\vskip 0.2in
\begin{center}
\centerline{\includegraphics[width=\textwidth]{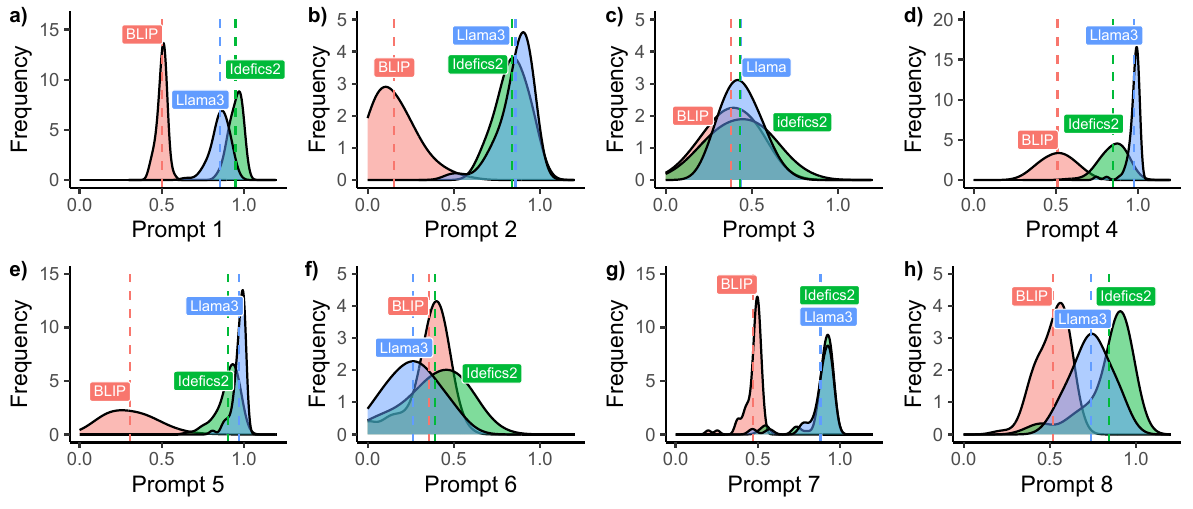}}
\caption{Proportion accuracy from \textbf{multimodal evaluations} of VLM performance on spatial recognition prompts (see Table \ref{table-prompts}) in which each prompt was paired with an image from \textsc{TableTest}. Dashed lines in each panel depict overall accuracies of each VLM and density plots depict performance distributions across \textsc{TableTest}'s 64 objects, as organized by whether the object served as ``object A'' in prompt templates and variations. Humanlike performance anticipated to be at ceiling (accuracy = 1.0); VLMs with humanlike performance will exhibit density patterns akin to that shown in Figure \ref{figure-textonly}g below.}
\label{figure-multimodal}
\end{center}
\vskip -0.2in
\end{figure*}

For every evaluation, we used prompts and their variations with every 2-object image in \textsc{TableTest} as VLM queries. Table \ref{table-overallperformance} and Figure \ref{figure-multimodal} present their overall results. To assess whether performance differences were reliably different from one another, we pooled results by the leftmost object in each image (which we refer to as \emph{object A}). For example, any image in which the leftmost object was a blender (\emph{blender}-cupcake, \emph{blender}-toaster, \emph{blender}-banana, etc.) was treated as a \emph{blender} image, and we pooled performance results over all \emph{blender} images, \emph{toaster} images, and so on. We subjected those pooled data to nonparametric statistical analysis, because such tests eschew unnecessary assumptions about performance distributions, and are accordingly more conservative than alternative analyses. For brevity we report on only relevant main effects and interactions; we make ancillary analyses available online.

\subsubsection{Prompt 1}

Prompt 1 queried for true/false judgments of a given spatial relation (see Table \ref{table-prompts}). We evaluated the prompt by varying whether the spatial relation was \emph{left} or \emph{right} and whether the order of the objects matched or mismatched the image (rendering the relation true or false, respectively). This yielded a total of 16,128 queries (4 variations $\times$ 4,032 images in \textsc{TableTest}) to each of the 3 VLMs. Chance performance was 50\% accuracy.

The accuracies of the generated responses significantly differed for each VLM (Friedman nonparametric ANOVA, $\chi^2 = 122.28$, $p < .001$): BLIP's responses were 50\% accurate, Llama3's responses were 86\% accurate, and Idefics2's were 95\% accurate (see Table \ref{table-overallperformance}, column 1; and Figure \ref{figure-multimodal}a). We ran analogous analyses for Prompts 2-8 below, and they were all significant, so we omit these results from further reporting (but include them in statistical scripts online).

VLMs produced more correct answers when relations used \emph{left} than \emph{right} (79\% vs. 75\% correct; Wilcoxon test, $z = 4.22$, $p < .001$), both in aggregate and for every individual VLM. Likewise, they were more accurate for true relations than false ones (88\% vs. 66\% correct; Wilcoxon test, $z = 6.54$, $p < .001$), in aggregate and for each individual VLM. In other words, VLMs were better at recognizing that a relation was present in an image than recognizing that it was absent. BLIP produced notable underperformance on this metric: its was 73\% accurate at recognizing which relations were true and only 23\% accurate at recognizing which were false.

\subsubsection{Prompt 2}

Prompt 2 directed VLMs to select a correct relation from four separate options (such as \emph{the blender is to the left of the cupcake}). We randomized the orders of those options for each query, and ran 4K queries (1 for each image) on each VLM. Table \ref{table-overallperformance}, column 2 shows overall performance by VLM. The highest performing VLM was Llama3 (86\% correct), which means that for 86\% of images, Llama3 selected 1 of the 2 correct options it was presented (see Figure \ref{figure-multimodal}b).

In addition to this accuracy measure, we analyzed for potential biases that the VLMs could have exhibited based on two metrics: 1) whether the VLM selected an option that used \emph{left} or \emph{right}, and 2) whether the VLM selected an option in the order [\emph{object A, object B}] or else [\emph{object B, object A}]. An unbiased VLM should not exhibit preferential treatment between these choices (i.e., it should select \emph{left} as often as \emph{right} and the order \emph{object A, object B} as often as its inverse). Analyses revealed biases that reliably differed for each VLM. VLMs selected \emph{left} on 7\% of trials (BLIP), 51\% of trials (Llama3), and 55\% of trials (Idefics2; Friedman nonparametric ANOVA, $\chi^2 = 94.17$, $p < .001$). Preferences for the order [\emph{object A, object B}] were: BLIP (16\%), Llama3 (15\%), and Idefics2 (51\%; Friedman nonparametric ANOVA, $\chi^2 = 76.55$, $p < .001$). We synthesize these results as follows: Llama3 selected corrected responses more often than not, but when it did so, it preferred options that listed the objects in the order [\emph{object B, object A}] instead of the reverse order. BLIP performed suboptimally, in part because it always preferred to select \emph{right} over \emph{left} and in the order [\emph{object B, object A}] regardless of whether the provided image depicted those relations. Idefics2 exhibited weak biases for these factors.

\subsubsection{Prompt 3}

Prompt 3 was similar in every respect to Prompt 2, except that it directed VLMs to select relations that don't hold in the provided image. An accurate response to the query for, e.g., Figure \ref{figure-tabletest}b would be to select one of the following responses: [\emph{the blender is to the right of the cupcake}; \emph{the cupcake is to the left of the blender}]. All three VLMs performed worse than chance (BLIP = 38\% correct; Llama3 = 44\% correct; Idefics2 = 43\% correct; see Figure \ref{figure-multimodal}c), meaning that all VLMs performed worse than chance. These results reveal systematic deficiencies in identifying relations that do not hold in an image.

One way to explain worse than chance performance is if VLMs exhibit various biases to select an option based on the structure rather than the semantic meanings of options. For instance, VLMs differentially preferred \emph{left} or \emph{right} in the options they selected: BLIP (17\% \emph{left}), Llama3 (55\% \emph{left}), and Idefics2 (79\% \emph{left}; Friedman nonparametric ANOVA, $\chi^2 = 127.51$, $p < .001$). All three VLMs preferred descriptions in the [\emph{object B, object A}] order over the [\emph{object A, object B}] order, but the extent to which they exhibited this preference differed drastically: BLIP (35\% [\emph{object A, object B}] order), Llama3 (4\%), Idefics2 (50\%). These results partially explain suboptimal performance.

\subsubsection{Prompt 4}

Prompts 1-3 used propositions that consisted of a single spatial relation -- either \emph{left} or \emph{right} but not both. This special linguistic context could have introduced biases, and so Prompt 4 used questions that presented both spatial relations in the context of a disjunction, e.g., \emph{to the left of or to the right of}. We varied the order of the objects and the order of the relations in the disjunction, yielding 16,128 queries (4 variations $\times$ 4,032 images in \textsc{TableTest}).

The VLMs varied widely in their performance (see Figure \ref{figure-multimodal}d): Llama3 performed at ceiling (98\% correct), and Idefics2 yielded 85\% correct responses. BLIP, however, performed no better than chance (51\% correct). In aggregate, the order of the relations in the disjunction did not affect performance (79\% accurate for \emph{to the left of or to the right of} and 77\% for \emph{to the right of or to the left of}; Wilcoxon test, $z = .93$, $p = .35$). Neither did the order of the objects ([\emph{object A, object B}] = 77\% correct; [\emph{object B, object A}] = 79\% correct; Wilcoxon test, $z = 1.30$, $p = .19$). These two factors reliably interacted with one another (Wilcoxon test, $z = 5.9$, $p < .001$;  see Table \ref{table-prompt45results} top rows), that is, \emph{to the left of or to the right of} was preferred for one order but not the other. Both BLIP and Idefics2 exhibited the interactive pattern, whereas Llama3 performed optimally under all Prompt 4 variations.

This interaction surprised us: one post-hoc hypothesis of it is that VLMs systematically erred by basing responses on the first relation in the disjunct -- hence for the second row in Table \ref{table-prompt45results}, they erroneously responded that object A was to the \emph{right} of object B because ``to the right of'' was the first option in the disjunct. This explanation predicts that VLMs should err in bizarre ways when introducing an irrelevant relation at the beginning of a disjunct. That is, the presence of this irrelevant relation should cause VLMs to erroneously use it in a response, and this pattern should occur more often when the relation appears at the beginning of a disjunct than at the end. Prompt 5 sought to test this prediction.

\subsubsection{Prompt 5}

\begin{table}[!b]
\centering
\begin{tabular}{lc} 
\hline
\textbf{Queries}           & \textbf{\%}  \\ 
\hline
\textsc{Prompt 4} & \\
\ \ Is \emph{A} to the left of, or to the right of \emph{B}? & 89\%     \\
\ \ Is \emph{A} to the right of, or to the left of \emph{B}? & 66\%     \\
\ \ Is \emph{B} to the left of, or to the right of \emph{A}? & 70\%     \\
\ \ Is \emph{B} to the right of, or to the left of \emph{A}? & 88\%     \\
\hline
\textsc{Prompt 5} & \\
\ \ Is \emph{A} to the left of, to the right of, or \textbf{inside of} \emph{B}? & 72\%     \\
\ \ Is \emph{A} \textbf{inside of}, to the right of, or to the left of \emph{B}? & 73\%     \\
\ \ Is \emph{B} \textbf{inside of}, to the left of, or to the right of \emph{A}? & 69\%     \\
\ \ Is \emph{B} to the right of, to the left of, or \textbf{inside of} \emph{A}? & 76\%     \\
\hline
\end{tabular}
\caption{Percentages of correct responses to Prompt 4 (top rows) and Prompt 5 (bottom rows) as a function of the four different prompt variations presented in each query. Bolded text highlights placement of an irrelevant relation in the disjunctive set of relations presented (see text).}
\label{table-prompt45results}
\end{table}

Prompt 5 was similar in every respect to Prompt 4, except that it used introduced the irrelevant relation ``inside of'' to every disjunctive set of options presented to VLMs on every query (see Table \ref{table-prompt45results} bottom rows). As with the previous set of results, VLMs varied in their performance (see Figure \ref{figure-multimodal}e): Llama3 performed at ceiling (97\% correct), and Idefics2 yielded 90\% correct responses. BLIP performed worse than chance (31\% correct). Irrelevant relations were placed in an order designed to test whether the post-hoc explanation described was true. The patterns of responses produced by the VLMs suggest otherwise: the order of the relations in the disjunction yielded a pattern opposite to what the explanation predicts (74\% accurate when \emph{inside of} was the first disjunct and 71\%  when it was the last; Wilcoxon test, $z = 2.24$, $p = .03$). The order of the objects had no effect on performance ([\emph{object A, object B}] = 73\% correct; [\emph{object B, object A}] = 73\% correct; Wilcoxon test, $z < .001$, $p = .99$). And these two factors yielded no reliable interaction (Wilcoxon test, $z = .08$, $p = .94$, which again rules out the post-hoc explanation of the puzzling in interaction discovered for Prompt 4.

Llama3 performed optimally under all Prompt 5 variations. The other two VLMs exhibited reductions in performance as a result of the introduction of an irrelevant spatial relation, yielding overall reductions in accuracy from Prompt 4 to Prompt 5.

\subsubsection{Prompt 6}

Prompts 1-5 were all ``closed-form'' prompts, i.e., the information required to produce a meaningful response was contained within the prompt itself. Prompts 6-8 are all fill-in-the-blank, ``open-form'' prompts that demand adequate identification of objects along with the roles they play in spatial relations. Prompt 6 presented incomplete sentences such as: \emph{The [blank] is to the left of the [blank]} and asked VLMs to fill in the blanks according to the image provided. For example, if Figure \ref{figure-tabletest}b was provided as an input image, an appropriate response to the query above would be: \emph{blender, cupcake}, or else some semantic equivalent, e.g., \emph{appliance, food}. An inappropriate response could be because of any deviation from this format, e.g., if a response described 1 object, or else 3 or more objects, or else provided some summary of the image, and so on.

Prompt 6 presented two variations of incomplete sentences, i.e., the one above as well as the inverse spatial relation: \emph{The [blank] is to the right of the [blank]}. We paired each variation to one half of the images in \textsc{TableTest} and subjected all three VLMs to these 4,032 queries, yielding 12,096 queries in total. We coded responses as being accurate only if they responded appropriately as outlined above, and if they used the relevant Objaverse description of each object or some semantically related term (see Appendix \ref{appendixA}), because preliminary testing showed that all of the VLMs produced object labels that match those descriptions.

All VLMs performed poorly on this task (see \ref{figure-multimodal}f): Idefics2 performed best at 39\% accuracy, BLIP at 36\%, and Llama3 at 26\% of queries. The relation in each incomplete sentence affected aggregate performance: \emph{left} yielded 44\% accuracy and \emph{right} yielded 24\% accuracy (Wilcoxon test, $z = 5.3$, $p < .001$). This pattern was consistent for Idefics2 and Llama3, but BLIP produced \emph{29\%} accurate responses regardless of the relation.

\subsubsection{Prompt 7}

Prompt 6 probed for VLMs to fill in two separate blanks, and we coded responses as correct only if they described relations in the correct order. Poor performance may be attributed to this ordering constraint, and so Prompt 7 presented almost identical incomplete sentences with only one blank, e.g., \emph{The [blank] is on the left side of the table.} We varied only the relation in the prompt (\emph{left} or \emph{right}), and we coded the resulting open-ended data using a method analogous to Prompt 6.

VLM accuracy was markedly higher on this prompt: Idefics2 (89\%), Llama3 (88\%), BLIP (47\%). None, however, achieved human competence. The relation did not significantly affect performance (\emph{left} $=$ 76\% correct vs. \emph{right} = 73\% correct; Wilcoxon test, $z = .24$, $p = .81$), either in aggregate or for the individual VLMs.

\subsubsection{Prompt 8}

Prompt 8 tasked VLMs with filling in the blank for the preposition that completed the sentence, \emph{The blender is to the [blank] of the cupcake} or a similar sentence with the object names reversed. As before, we paired each type of sentence to half the \textsc{TableTest} images, yielding 4,032 queries per VLM. The task permitted straightforward response coding, because any response other than the terms ``left'' or ``right'' was incorrect. No VLM approached human competence on this task: Idefics2 produced 84\% correct responses, Llama3 74\%, and BLIP 52\% (see Figure \ref{figure-multimodal}h).

In aggregate, the order of the objects in a sentence affected performance: the [\emph{object A, object B}] yielded 81\% correct responses while the reverse order yielded 60\% correct responses (Wilcoxon test, $z = 5.7$, $p < .001$). Yet this pattern was not consistent across the different VLMs: BLIP was 92\% correct for [\emph{object A, object B}] and 12\% correct for its reverse; Llama3 performed equivalently for both orders; and Idefics2 was 76\% correct for [\emph{object A, object B}] and 93\% correct for its reverse. 

\subsubsection{Summary}

Multimodal evaluations revealed that the following factors reliably perturbed patterns of VLM spatial relation recognition: whether a relation was true or false in an input image (Prompts 1, 4); whether a description concerned the relation \emph{left} or \emph{right} (Prompts 2-4, 6); whether the objects in the relation were in the order [\emph{object A, object B}] or not (Prompts 3, 4, 6, 8); and the presence of an irrelevant spatial relation in the description (Prompt 5). 

None of these factors should have had a meaningful impact on the recognition of spatial relations in images.

\section{General Discussion}

Pick any image off the internet and answer these two questions about it:

\begin{enumerate}[nosep]
	\item ``Does the image depict a turtle to the left of a fox?''
	\item ``Does it depict a fox to the right of a turtle?''
\end{enumerate}

No matter the image, your answer to these two questions should be identical, because they probe for logically equivalent relations. Your answers are not a result of any profound ability to meditate and extract spatial information from imagery: they're trivial. You either perceived those relations, or you didn't. And vision-language models (VLMs) designed to mimic the visual reasoning abilities of humans should, at a minimum, reproduce this trivial behavior reliably \cite{blaha2022understanding}. Few such analyses of VLM reliability exist, so we developed a novel evaluation protocol designed to test for it, and found that VLMs routinely produce incoherence when probed for relations in images. They yielded stark differences in performance that depended on nearly every variable we manipulated. The different prompts, trivial variations of those prompts, and even different combinations of objects (we omitted such analyses for brevity) could all perturb VLM performance. Yet even these perturbations were not reliable from one prompt to the next.

Because we sampled VLM performance from disparate architectures, the results suggest the potential for widespread unreliability in any system that uses similar training datasets. Systematic differences between architectures can be explained by numerous factors, but as \citet{kamath2023s} note, most if not all robust VLMs are trained on large-scale image caption corpora, and image captions rarely describe spatial relations. When they do, they describe one or two relations, but they never describe all the relations present in an image, because such captions are likely to be useless for human consumption. Because of these training limitations, architectures may be vulnerable to generating unreliable responses for many kinds of spatial relations (e.g., \emph{inside, on, above, below, between}) beyond the ones we studied. Yet the investigations also suggest opportunities for augmenting existing corpora by making use of trivial variations, which could yield more efficient and accurate training regimes.

These results do not impugn the use of VLMs in general. The three under evaluation are all reliable at many image- and text-processing tasks, and indeed, both Llama3 and Idefics2 produced highly accurate, humanlike responses for at least 2 of the 8 spatial relation recognition prompts we used. The results suggest instead that researchers should use caution when equating a model's competence on any particular prompt with its overall competency for a particular cognitive task. Even if a VLM responded accurately to one prompt, it's not the case that it can serve as a general spatial reasoning engine, because such an engine should constructing coherent representations of space given similar inputs, and thereby respond sensibly to different kinds of prompts. For any VLM, it may be possible to train out low performance on a single prompt through fine-tuning, but unless the VLM constructs coherent representations, fine-tuning is unlikely to transfer to different prompts and trivial variations of them.

As Table \ref{table-overallperformance} shows, performance was much higher on text-only contexts than multimodal contexts. We used text-only contexts to serve as a control condition for comparison purposes only, and so this high level of accuracy does not suggest competence at verbal spatial reasoning. The text-only contexts we used are unlikely to occur in everyday usage, because they fully specify the spatial properties of each object. The VLMs performed at ceiling, but their performance dropped for Prompts 2 and 3. As Appendix \ref{appendixA} shows, trivial variations affect performance even for prompts such as these, e.g., reversing the order of sentences in text-only descriptions can perturb accuracy. These results call for further investigation of the reliability of text-only spatial relation processing.

This research was not meant as a survey of all VLMs, which may be infeasible given how frequently researchers update and release models. Yet it may serve as a guide for model designers themselves, who seldom test for systematic comprehension in the manner we describe above. Put differently, it may be necessary for VLM engineers to conduct benchmarking analyses such as the one we devised to show that their system performs reliably on a wide variety of tasks. And reliability across multiple prompts may indeed be a better metric of the power of VLM than performance on a single prompt, because the errors produced by a VLM that is highly reliable may be easier to mitigate than one that exhibits high performance on a single prompt but unreliable performance on trivial variations. A corollary of this point is that interventions designed to eliminate biases and perturbations should be reliable too, i.e., they should transfer to different kinds of prompts.  

In sum, none of the models we tested revealed consistent and reliable spatial relation comprehension, either from visual imagery or text. These results suggest limitations in how reliable such models are at higher-order spatial reasoning, which depends on the reliable recognition and processing of spatial relations. If advances yield VLM architectures the exhibit higher reliability, designers will need to engage in investigations such as the one we conducted with \textsc{TableTest} above, i.e., those that test across multiple prompts using imagery in which the spatial properties of the objects are known in advance.

\bibliographystyle{ieeenat_fullname}
\bibliography{sllm_paper}

\newpage
\onecolumn

\appendix

\section{Assessing accuracy for Prompts 6-8} \label{appendixA}

Due to the open-ended format of Prompts 6-8, we needed to develop a more robust method to measure accuracy. For instance, the raw responses that VLMs generated often used semantically related terms for objects, either by generating a synonym (e.g., \emph{cup} instead of \emph{mug}), a superordinate category (e.g., \emph{bottle} instead of \emph{wine bottle}), or else by misidentifying an object in a way that was reasonable (e.g., identifying a \emph{briefcase} as a \emph{lunch box}). We sought to develop a liberal accuracy metric that permitted such variations in responses. To find semantic alternatives, the raw responses for Prompts 6 and 7 were analyzed for each individual object. If a VLM routinely generated a semantic alternative in place of an expected description 5+ times, that alternative was entered into a lookup table. If a response described an object or its semantic alternative, the object name was replaced by a placeholder, and that placeholder was used to assess accuracy. Hence, a response was considered correct if it identified the two objects in the image or produced any acceptable variation of those two objects; and if the noun phrases used to describe those objects were found in the correct order. This metric counts a response as incorrect only if it fails to recognize the relation in the image.

On some responses to Prompt 6, VLMs generated full sentence descriptions instead of performing the fill-in-the-blank task as requested. We assessed accuracy using a modified version of the metric above that also took into account that full sentences could use spatial relations that weren't in the prompt itself. Hence, if an image depicted a wine bottle to the left of a mug, and the prompt asked for the VLM to complete this sentence: ``the [blank] is to the left of the [blank]'', then a response such as \emph{the cup is to the right of the bottle} was considered correct.

\section{Text-only context evaluations} \label{appendixB}

\noindent Text-only evaluations of Prompts 1-8 were analogous in almost every respect to multimodal evaluations, except that images in \textsc{TableTest} were replaced by a two-sentence vignette that described how the objects were placed on a table (see \S\ref{text-only-example}). For each evaluation, we varied whether the vignette described the left-side or the right-side object first.

\begin{figure*}[hb]
\vskip 0.2in
\begin{center}
\centerline{\includegraphics[width=\textwidth]{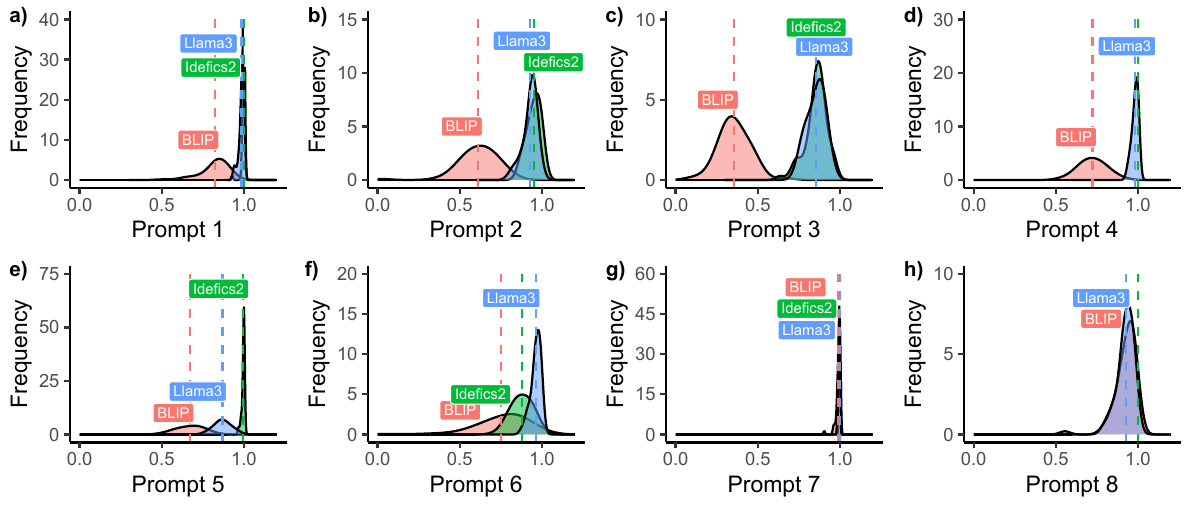}}
\caption{Proportion accuracy from \textbf{text-only evaluations} of VLM performance on spatial recognition prompts (see Table \ref{table-prompts}) in which each prompt was paired with text that described a particular image from \textsc{TableTest}. Dashed lines in each panel depict overall accuracies of each VLM and density plots depict performance distributions across \textsc{TableTest}'s 64 objects, as organized by whether the object served as ``object A'' in prompt templates and variations. Humanlike performance anticipated to be at ceiling (accuracy = 1.0).}
\label{figure-textonly}
\end{center}
\vskip -0.2in
\end{figure*}

\subsection{Prompt 1}

Prompt 1 queried for true/false judgments of a given spatial relation (see Table \ref{table-prompts}). Response accuracies differed reliably for each VLM, though two performed at ceiling: Llama3 and Idefics2's responses were 99\% accurate, while BLIP's were 82\% accurate (see Table \ref{table-overallperformance}, column 1; and Figure \ref{figure-textonly}a).

VLM performance didn't differ by whether the left-side or right-side object was described first (94\% in both conditions; Wilcoxon test, $z = 4.22$, $p < .001$), or whether the queried relation was \emph{left} or \emph{right} (Wilcoxon test, $z = 1.66$, $p = .09$). However, the truth of the queried relation affected performance: answers were more accurate for true relations than false ones (96\% vs. 90\% correct; Wilcoxon test, $z = 4.53$, $p < .001$).

\subsection{Prompt 2}

Prompt 2 queried VLMs to select a correct response from four options. The highest performing VLM was Idefics2 (95\% correct), followed by Llama3 (93\% correct) and BLIP (61\% correct; see Figure \ref{figure-textonly}b). 

Performance didn't differ significantly as a function of the order of relations in vignettes (83\% correct for both; Wilcoxon test, $z = .93$, $p = .35$). Their bias towards \emph{left} response differed by VLM: BLIP selected \emph{left} responses 36\% of the time, while Idefics2 and Llama3 selected them 61\% of the time (Friedman nonparametric ANOVA, $\chi^2 = 73.15$, $p < .001$). Preferences for the order [\emph{object A, object B}] were: BLIP (47\%), Llama3 (56\%), and Idefics2 (54\%; Friedman nonparametric ANOVA, $\chi^2 = 6.53$, $p = .04$), indicating a measurable but relatively inconsequential ordering bias.

\subsection{Prompt 3}

Prompt 3 directed VLMs to select relations that don't hold in the provided image. None of the three VLMs achieved humanlike competence (Idefics2 = 86\% correct; Llama3 = 85\% correct; BLIP = 36\% correct; see Figure \ref{figure-textonly}c), possibly reflecting general difficulty processing negative instructions.

The order of relations in vignettes significantly affected performance: VLMs were more accurate when the left-side object was described first vs. second (73\% vs. 65\%; Wilcoxon test, $z = 5.95$, $p < .001$). Only one VLM (BLIP) systematically showed a bias for choosing \emph{right} over \emph{left}; it did so on 63\% of trials. The other two VLMs showed no such bias (51\% \emph{left} for both; Friedman nonparametric ANOVA between VLMs, $\chi^2 = 70.23$, $p < .001$). Likewise, only one VLM (Idefics2) showed an ordering preference for the [\emph{object B, object A}] order (56\% of the time) over the [\emph{object A, object B}] order, while the other two VLMs showed no such bias (Friedman nonparametric ANOVA between VLMs, $\chi^2 = 70.23$, $p < .001$).

\subsection{Prompt 4}

Prompt 4 provided VLMs with a disjunctive description as a guide for responses (e.g., \emph{to the left of or to the right of}); it varied the order of the objects, and we assessed whether responses were accurate and whether they exhibited any biases. Accuracy was at ceiling for two models (98\% and 100\% accurate for Llama3 and Idefics2, respectively) and lower for BLIP (72\% accurate; see Figure \ref{figure-textonly}d). The order of objects, however, affected performance in aggregate ([\emph{object A, object B}] = 95\% correct; [\emph{object B, object A}] = 86\% correct; Wilcoxon test, $z = 6.96$, $p < .001$), and for the VLMs whose responses showed variation in accuracy. Likewise, the order of relations in the disjunction affected performance (94\% accurate for \emph{to the left of or to the right of} and 87\% for \emph{to the right of or to the left of}; Wilcoxon test, $z = 6.84$, $p < .001$).

\subsection{Prompt 5}

Prompt 5 was identical to Prompt 4, but it introduced an irrelevant relation either to the beginning or the end of the set of disjunctions provided in each query. This change lowered overall accuracy (Idefics2 = 99\% correct; Llama3 = 87\% correct; BLIP = 67\% correct; see Figure \ref{figure-textonly}e). It also reversed ordering bias, i.e., ([\emph{object A, object B}] = 82\% correct; [\emph{object B, object A}] = 87\% correct; Wilcoxon test, $z = 5.12$, $p < .001$). Likewise, the order of relations in the disjunction once again affected performance (89\% accurate when \emph{inside of} came first and 80\% when \emph{inside of} came last; Wilcoxon test, $z = 6.61$, $p < .001$).

\subsection{Prompt 6}

While Prompts 6-8 were open-format in nature, they were easier to code than multimodal prompts (see Appendix \ref{appendixA}) because they required no image processing or object detection and labeling: vignettes stipulated the correct names of objects, and VLMs had only to use these names in their appropriate appropriate relational context. Prompt 6 called for VLMs to fill in two blanks in an incomplete sentence. Only Llama3 achieved humanlike competence even in these circumstances (BLIP = 75\% correct; Idefics2 = 88\% correct; Llama3 = 96\% correct; see Figure \ref{figure-textonly}f). VLMs did not reveal any relational preference, either in aggregate or individually (\emph{left} = 87\% correct; \emph{right} = 86\% correct; Wilcoxon test, $z = 1.62$, $p = .11$).

\subsection{Prompt 7}

Performance on Prompt 7 (which asked VLMs to identify one object in isolation and place it on the appropriate side of the table) was at ceiling: all VLMs achieved $>$ 99\% accuracy (Figure \ref{figure-textonly}g). Both the prompt itself and the vignette describing objects on the table used identical language, and so this prompt was an instance of text-matching and did not test any fundamental competence in spatial cognition. 

\subsection{Prompt 8}

Prompt 8 was similar to Prompt 7: it asked for VLMs to fill in the missing relation in an incomplete sentence, and it did not require object identification. Performance dropped in comparison: all three VLMs showed humanlike competence, but at a rate lower than Prompt 7 (BLIP = 93\% correct; Llama3 = 93\% correct; Idefics = 100\% correct). And responses yielded a detectable bias towards better performance for the ([\emph{object B, object A}] order (97\% correct) vs. the reverse (93\% correct; Wilcoxon test, $z = 4.30$, $p < .001$).

\end{document}